\documentclass[runningheads]{llncs}
\usepackage[colorlinks=true, urlcolor=blue, citecolor=blue, filecolor=blue, anchorcolor=blue, linkcolor=blue]{hyperref}

\usepackage{siunitx}
\usepackage{graphicx}

\newcolumntype{C}[1]{>{\centering\let\newline\\\arraybackslash\hspace{0pt}}m{#1}}

\begin{document}
\title{Predicting Clinical Diagnosis from Patients Electronic Health Records Using BERT-based Neural Networks}
\titlerunning{Predicting Clinical Diagnosis from Patients EHRs Using BERT-based NN}
\author{Pavel Blinov\inst{1} \and
Manvel Avetisian\inst{1} \and
Vladimir Kokh\inst{1} \and
Dmitry Umerenkov\inst{1} \and
Alexander Tuzhilin\inst{2}}
\authorrunning{P. Blinov et al.}
%
\institute{Sberbank Artificial Intelligence Laboratory, Moscow, Russia \\
\email{\{Blinov.P.D,~Avetisyan.M.S,~Kokh.V.N,~Umerenkov.D.E\}@sberbank.ru} \\
\and New York University, New York City, USA \\
\email{atuzhili@stern.nyu.edu}}

\maketitle              
\begin{abstract}
In this paper we study the problem of predicting clinical diagnoses from textual Electronic Health Records (EHR) data. We show the importance of this problem in medical community and present comprehensive historical review of the problem and proposed methods. As the main scientific contributions we present a modification of Bidirectional Encoder Representations from Transformers (BERT) model for sequence classification that implements a novel way of Fully-Connected (FC) layer composition and a BERT model pretrained only on domain data. To empirically validate our model, we use a large-scale Russian EHR dataset consisting of about 4 million unique patient visits. This is the largest such study for the Russian language and one of the largest globally. We performed a number of comparative experiments with other text representation models on the task of multiclass classification for 265 disease subset of ICD-10. The experiments demonstrate improved performance of our models compared to other baselines, including a fine-tuned Russian BERT (RuBERT) variant. We also show comparable performance of our model with a panel of experienced medical experts. This allows us to hope that implementation of this system will reduce misdiagnosis.

\end{abstract}
\section{Introduction} \label{Introduction}
The process of digital transformation in medicine has been going for a while, providing faster and better treatment results in many cases through the use of modern computer science and Artificial Intelligence (AI) methods~\cite{ref1}. Digitization and subsequent analysis of medical records constitutes one such area of digital transformation that aims to collect broad types of medical information about a patient in the form of EHR, including digital measurements (laboratory results), verbal descriptions (symptoms and notes, life and disease anamnesis), images (X-Ray, CT and MRI scans, etc.) and document the treatment process of a patient.

In this paper, we focus on the analysis of EHR with the purpose of providing clinical decision support by predicting most probable diagnoses during a patient's visit to a doctor. This problem is complicated by abundance of large volumes of structured and unstructured medical information stored across multiple systems in different data formats that are often incompatible across these systems. Although there exists an emerging FHIR standard (Fast Healthcare Interoperability Resources) for the EHR data~\cite{ref2} the goal of which is to unify the process of storing and exchanging medical information, unfortunately, very few existing Hospital Information Systems (HISes) support it. All this complicates the task of diagnosis prediction based on the EHRs since many of them contain extensive amounts of unstructured, poorly organized and "dirty" data that is less amenable to the analysis using the AI-based methods, unless this data is cleaned and preprocessed appropriately.

Providing clinical decision support in diagnosis prediction during a patient's visit to a doctor is important because many patient's visits, in fact up to 30\% in the US, are misdiagnosed~\cite{ref6}. This is also true in some other countries~\cite{ref7}. We formulate the aforementioned clinical decision support problem as a multi-label text classification of clinical notes (anamnesis and stated symptoms) during a patient visit, where the classification is performed for a wide range of diagnosis codes represented by the International Statistical Classification of Diseases (ICD-10)~\cite{icd10}.

In this paper, we make the following contributions. First, we propose a novel BERT-based model for classification of textual clinical notes, called \emph{RuPool-BERT}, that differs from the previously proposed models by the way of the FC-layer composition that is described in Section \ref{PropModels}. Second, we compare the performance of our method with various baselines across different text representation techniques and classification models. Third, we compare the performance of the BERT model pretrained on a large corpus of out-of-domain data~\cite{ref20} with the BERT model pretrained exclusively on in-domain data and using an in-domain tokenizer. Finally, we demonstrate the advantage of the proposed models and their comparable results with a human baseline in Section~\ref{Experiments}.

It is important to note that the clinical decision support system described in the paper will \emph{not} serve as a doctor's replacement but, rather, constitutes an unbiased intelligent diagnosis generator and, therefore, should only \emph{assist} the doctors in their diagnostic decisions.

\section{Related Work}
There have been many approaches to the analysis of the EHR-data and predicting the diagnosis codes (ICDs) proposed in the literature that are related to our work. In particular, papers~\cite{ref13,ref14} address the task of diagnoses prediction by using the entire patient history. In our study we explicitly do not take history into account because at the current stage of our project we focus on isolated visits; however, the history is partially accessible for the model due to the availability of the anamnesis field. The level of granularity for ICD-code in paper~\cite{ref13} is quite similar with ours. We also mainly operate on the second level of ICD-10 classification code hierarchy, but intentionally restrict the number of classes up to 265 (see details in the Data Section~\ref{Data}).

The authors of~\cite{ref15} developed a Deep Neural Network-based (DNN) diagnosis prediction algorithm, but only for the pediatric EHRs. In contrast to this, we do not restrict ourselves to any age specifications. Instead, our training set fully covers all the age groups, including having about 20\% of the training data visits being children under the age of 14 years. The papers~\cite{ref14,ref15,ref8} proposed methods based on the DNN (mostly its recurrent variants), while in this paper we focus on more recent state-of-the-art transformer-based neural architectures.

In the recent couple of years a new class of neural architectures called transformers were developed, which allow to significantly improve performance on the whole range of NLP tasks (e.~q. question answering, named entity recognition, sentiment classification etc.). The most known member of this family is the language representation model called Bidirectional Encoder Representations from Transformers (BERT)~\cite{ref16}. Note that the BERT-like models have been applied to the EHR-data before~\cite{ref17,ref18}. In particular, \cite{ref17} presents a system of assigning ICD-10 codes to non-technical textual summaries of medical experiments. The original experiment dataset (about 8,000 samples) was in German; but the authors achieved significant performance improvement (more than 6\% for the F1-measure) by translating it into English and then applying the BERT model to the translated EHRs. This shows, among other things, that each language has its own linguistic and cultural idiosyncrasies and that there are ‘easier’ vs. 'harder' languages for machine learning models. In this paper we do not use such a "translation trick" and work directly with the original language.

Fei Li et al.~\cite{ref18} investigated the problem of BERT model fine-tuning for biomedical and clinical entity normalization. The training EHR notes used in that paper have millions of entries and in that way are similar to our study in its ability to handle large EHR datasets. Another significant advantage of this paper is the comparative analysis between pretrained general and biological domain BERT models. In our work, we also use a general Russian pretrained BERT model~\cite{ref20}. Since the lexical and the syntactic structure of a special domain languages can be very different from the general one, we experimented with an in-domain tokenizer and the model trained from scratch on the Russian EHR data.

Most of the EHR research focuses on the English language based electronic health records. Even~\cite{ref17} translated their health records from German into English to leverage the power of the previously conducted English-based research. In contrast to this, we think that it is crucial to successfully apply artificial intelligence and NLP-based methods to other languages, which is done only occasionally. For example, the amount of the EHR-related studies is very limited for the Russian language. In fact we could find only one such recent paper~\cite{ref12}, where the authors studied a related classification problem, but only for four ICD codes (D50, E11, E74, E78) and on a small dataset having about 8,000 cases where they used a gradient boosting algorithm on a laboratory tests data as input. 

In this paper, we follow this principle and analyze electronic health records of medical patients written in Russian. Unlike several other studies conducted on medium-size EHR datasets, such as the Medical Information Mart for Intensive Care (MIMIC) dataset~\cite{ref11} containing about 60,000 intensive care unit admissions~\cite{ref8,ref9,ref10}, we work with a large dataset of about 4,000,000 patients' visits to various clinics in Russia. 

Finally, the authors in ~\cite{ref18} and ~\cite{ref20} use the conventional BERT method (and use the output of the classification token (CLS) for the upstream tasks). In this paper we show that this is not an optimal solution since the performance of the BERT model can be further improved by the extension of this layer. Therefore, we venture beyond the conventional BERT architecture and propose a modification to it that we call RuPool-BERT. Furthermore, we show that this extension outperforms the conventional BERT approaches for our classification problem.

\section{Data} \label{Data}
In this project, we worked with three real-world anonymized datasets containing information about patients' visits to the networks of clinics in Russia. The first two datasets (we call them DataN and DataM in the paper) pertain to two large private networks of clinics and the third one (we call it DataT) pertains to the network of public clinics. We used only the symptom and anamnesis fields for each patient visit since all other fields where substantially different across the datasets. All the relevant data was concatenated into a single textual content field (since it was initially stored in different formats across different datasets). We did not apply any special preprocessing to this field, and therefore it was presented to the model as a raw text including typos, abbreviations and misspellings. The main statistics for each dataset are summarized in Table~\ref{tab1}.

\begin{table}
\centering
\caption{Statistics of the datasets.}\label{tab1}
\begin{tabular}{|C{1.5cm}|C{2.3cm}|C{1.3cm}|C{1.5cm}|C{1.9cm}|C{1.3cm}|C{1.3cm}|}
\hline
Dataset name & Split & \# of patients & \# of visits & Avg \# of visits per patient & From & To \\
\hline
DataN & train/validation & 251,763 & 1,685,253 & 6.69 & 2005-01 & 2018-12 \\
DataM & train/validation & 177,715 & 563,106 & 3.17 & 2013-01 & 2019-06 \\
DataT & test & 694,063 & 1,728,529 & 2.5 & 2014-01 & 2019-10 \\
\hline
\end{tabular}
\end{table}

As Table~\ref{tab1} demonstrates, the three datasets collectively have 3,976,888 visits of 1,123,541 patients over the period of almost 15-years. To the best of our knowledge, this is the largest such study (in terms of the number of cases and time duration) for the Russian language and one of the largest globally\footnote{Although the data does not contain any personal information, we cannot publicly release it due to certain legal restrictions.}.

Patient visits were split into train, validation and test sets. The whole DataT part was assigned to the test set, and the union of DataN and DataM sets was randomly split in the 80/20 proportion to make the train and the validation sets. The final cardinalities of the train, validation and test sets were 1,798,687 (45.23\%), 449,672 (11.3\%) and 1,728,529 (43.47\%) respectively. The validation set was used exclusively to fine-tune hyperparameters for the baseline BERT model. The test set was used to compare different baselines with the proposed models. The reason we decided to keep the entire DataT only as the test set lies in the more reliable nature of this data. More specifically, for each visit in DataT, we have an confirmed diagnosis.

The full spectrum of ICD-10 consists of 71,932 codes arranged in a hierarchical manner. A single code can be represented by 3 to 7 characters depending on the level of disease specification. Internally, an ICD code has 3 part structure. The first part, up to the dot, represents a distinct disease. For example, D30 is the code for the neoplasm of urinary organs. After the dot follows potential specifying elements (e.~g. D30.0 represents neoplasm of a kidney, D30.01~–~neoplasm of the right kidney, D30.02~–~neoplasm of the left kidney, etc.).

We selected \(K=265\) categories of codes for this study because such number of codes (diseases) is enough to cover up to 95\% of all the cases in the training set. As with many other natural distributions, distribution of diagnoses is very skewed, i.e., the first 19 codes accounting for 50\% of all the cases in the training set (with J06~–~11.5\%, I11~–~7.4\%, E11~–~4.3\%, M42~–~3.9\% and so on up to D72~–~0.03\% and L40~–~0.03\%). We have also experimented with \(K=1,000\) codes. This selection of codes were obtained from the same 265 codes by extending them with available second ICD parts (e.g. J06.0, J06.8, J06.9, etc.) and selecting the most frequent 1,000 of them. This second case of \emph{K = 1000} codes significantly complicates the classification task since it requires to predict more fine-grained diagnoses.

\section{Proposed Models} \label{PropModels}
\begin{figure}
\includegraphics[width=\textwidth]{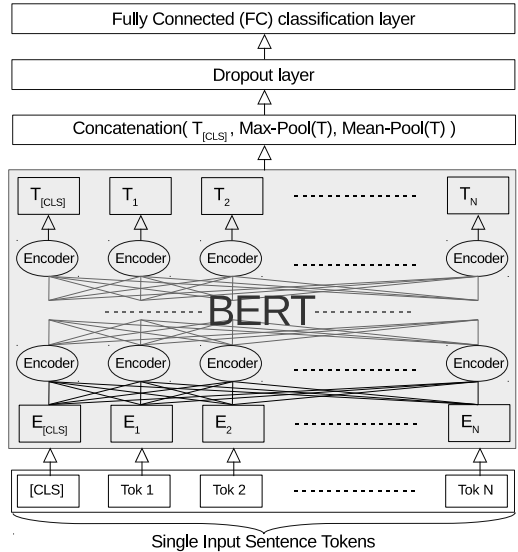}
\caption{Architecture of RuPool-BERT model.} \label{fig1}
\end{figure}

In this section we present our classification model that predicts the 265 ICD codes based only on textual information of the patient visits to the clinics. We solve the classification problem by using a transformer-based BERT model~\cite{ref16} architecture presented in Fig.~\ref{fig1} (referred as RuPool-BERT hereafter). The inputs to the model constitute text from the symptoms and the anamnesis fields of the patients visit to a doctor that are concatenated into a single text sequence. Most of the transformer-based models use the approach proposed in~\cite{ref19} to represent raw input sequences in terms of sub-words (tokens) and keep balance between character and word information. That allows to naturally process out of vocabulary words (typos, misspellings, etc.). The distribution of sequence lengths (for the train set) showed that the mean number of tokens is around 79 and median value is about 57. Therefore, we decided to allow some margin by limiting each sequence to \(N=128\) tokens.

Each input token is represented by learnable H-dimensional vector called \emph{embedding} (which in turn is the sum of three embeddings: token, segment and positional). Therefore, the text tokenizer and the trained model are linked together. As a base tokenizer (with the vocabulary of approximately 120k tokens) and a model, we used RuBERT (architecturally the same as base BERT model for the English: 12 transformer block layers, hidden size H=768 and 12 self-attention heads)~\cite{ref20} because it significantly outperformed the multilingual variant of BERT, as was shown in~\cite{ref20}.

The gray part of the model presented in Figure~\ref{fig1} is the same as in ~\cite{ref16}, which methodically explains this part of the network architecture and the process of pretraining. The authors of~\cite{ref16} specifically designed the CLS token as classification one, and the linear fully-connected layer is added to the last hidden state \(C\): \(T\textsubscript{[CLS]} \in R\textsuperscript{H}\). We propose to concatenate this state with two additional parts, namely max and mean pooling over the whole last encoder states \emph{T\textsubscript{i}} along the sequence dimension. Both operations also return embeddings from \(R\textsuperscript{H}\). With a new hidden state vector \(C \in R\textsuperscript{3H}\) and fully-connected classification layer with weights \(W \in R\textsuperscript{K3H}\), where \(K=265\) – number of ICD codes, the diseases probabilities after applying sofmax function is then \(P=softmax(CW\textsuperscript{T})\). To prevent overfitting, the dropout operation was applied to the \(C\) layer.

To fully leverage the difference in vocabulary between general texts and medical records, we trained a tokenizer with a vocabulary of 40k tokens on all the texts from DataN dataset. The tokenizer is identical to the one used in the original BERT model with the difference coming from the training data word distribution. Our expectation was that such a medical-domain tokenizer will allow the model to capture a wider range of medical linguistics phenomena.
Using this tokenizer, we pretrained the BERT model with masked language modelling task on data from DataN dataset containing about 1.7 million records of patients visits, which took us about 2 weeks on a Tesla K40 GPU.
This pretrained model with a standard pooling scheme (referred RuEHR-BERT hereafter) was finetuned for disease classification. RuEHR-BERT can be directly compared with the RuBERT model which has the same architecture but a tokenizer trained on general Russian language texts.

\section{Experiments} \label{Experiments}
To evaluate the performance of the proposed model, we compared it with the following baselines: an RNN model (with GRU units), the FastText model~\cite{ref21} and the multilingual Universal Sentence Encoder (USE)~\cite{ref22}. We focused on these baselines because it was feasible to do the direct comparisons with them, while comparing our method with several others was either impossible or very hard to achieve due to accessibility, language incompatibility and other practical reasons.

To remove the effect of hyperparameter tuning and compare all the BERT models under the same conditions, hyperparameters were kept the same across all these models. First, we find the best set of parameters (with respect to the validation data) for the baseline RuBERT model and used them thereafter. We report our results after 5 training epochs with \(batch~size = 128\), optimizer being AdamW and the starting learning rate of \num{3e-5} with the Binary Cross Entropy loss. By fine-tuning the hyperparameters for our proposed models on the validation set the results could be improved further.

We considered the following performance metrics in our study: the macro and weighted variants of the F1-measure~\cite{Manning} and such ranking measures as Mean Reciprocal Rank (MRR) and Hit@k~\cite{Manning} (i.e., Hit@1, Hit@3, Hit@5 and Hit@10). In cooperation with medical experts we empirically selected Hit@3 as our primary metric.

Table~\ref{tab2} summarizes the performance results of all the considered models on the test set. Note that a large performance gap between the \(F1\textsubscript{macro}\) and \(F1\textsubscript{weighted}\) measures across all the baselines reflects great disbalance in the class distribution (that was discussed in Section~\ref{Data}). The significant increase in \(F1\textsubscript{macro}\) for the proposed models (4\%) combined with increase in \(F1\textsubscript{weighted}\) (1\%) compared to baseline RuBERT model show that our proposed models performance gains are due to better classification of less frequent disease classes.

Also, we tried to train the 1,000 class ICD code classifier for our RuPool-BERT model (denoted as "RuPool-BERT 1k" in Table~\ref{tab2}). Note that, although the number of classes increased almost by the factor of four for the RuPool-BERT 1k model, the performance measured by the Hit@3 metric degrades significantly slower (42.97 vs. 70.14). As Table~\ref{tab2} demonstrates, RuPool-BERT 265 and RuEHR-BERT 265 show better performance than other tested models in terms of Hit@k and other measures.

\begin{table}
\centering
\caption{Models performance (\%) on the test set.}\label{tab2}
\begin{tabular}{|l|c|c|c|c|c|c|c|}
\hline
Model name & F1\textsubscript{macro} & F1\textsubscript{weighted} & MRR & Hit@1 & Hit@3 & Hit@5 & Hit@10 \\
\hline
USE & 19.54 & 39.59 & 54.69 & 41.25 & 62.62 & 70.66 & 79.89 \\
RNN & 24.33 & 43.34 & 58.02 & 44.17 & 66.97 & 74.79 & 83.37 \\
fastText & 24.49 & 44.19 & 59.00 & 45.27 & 67.81 & 75.84 & 84.54 \\
RNN+FastText & 25.44 & 44.00 & 59.11 & 45.05 & 68.43 & 76.22 & 84.71 \\
RuBERT 265 & 25.78 & 46.34 & 60.60 & 47.29 & 69.50 & 76.96 & 84.79 \\
\hline
RuPool-BERT 265 & 29.83 & 47.13 & 61.04 & 47.54 & 70.14 & 77.53 & 85.49 \\
RuEHR-BERT 265 & 28.61 & 46.84 & 61.01 & 47.51 & 70.00 & 77.61 & 87.76 \\
RuPool-BERT 1k & 8.95 & 24.03 & 37.13 & 25.94 & 42.97 & 50.51 & 59.46 \\
\hline
\end{tabular}
\end{table}

\begin{figure}
\includegraphics[width=\textwidth]{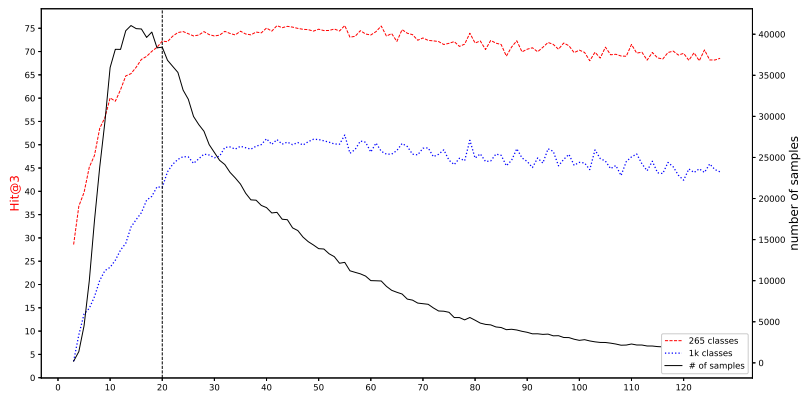}
\caption{Distribution of number of samples in the test set (solid black) and RuPool-BERT model Hit@3 metric (\%) (dashed red for 265 and dotted blue for 1k classes).} \label{fig2}
\end{figure}

Furthermore, we studied the dependence between the input text (symptoms and anamnesis) lengths and the Hit@3 metric by computing the metric for clinical notes with different input lengths. The results are presented in Figure~\ref{fig2}, where the text length is plotted on the x-axis and the number of test samples on the y-axis (for the black solid curve). Moreover, the red dashed and the blue dotted curves in Figure~\ref{fig2} show how the Hit@3 metric depends on the text length for the 265- and 1k-classification models respectively.

We can conclude from Figure~\ref{fig2} that the most reliable results of the model can be obtained for the cases where the input text sequence has at least 20 tokens, which constitutes 72.25\% of all the test visits. This is an important observation because it indicates when the results of our model can be shared with the doctors in practical clinical settings in the form of the "second opinion", i.e. it can be done only for the more extensively documented cases having at least 20 tokens.

Additionally, we compared the performance of our proposed models with experienced physicians in the diagnosis task described in Section~\ref{Introduction}. In particular, we filtered out test set according to above observation and randomly selected 530 visits (\(K*2=265*2\)) for the further markup process. We invited a panel of 7 medical experts (each with more than 10 years of practical experience) to participate in this study. Each of them was presented with these 530 visits and asked to point out up to 3 appropriate ICD codes (from \(K=265\)) for each case or reject it if the information to make a decision is insufficient. For the sake of a fair performance comparison, we asked the clinicians to make the decisions exclusively based on the EHR data (text from anamnesis and symptom fields) without any personal communications with the patients or any other parties. To assess reliability of agreement between the raters we computed Fleiss' kappa coefficient (among 1st place answers) \(k=0.37\) which corresponds to fair strength of agreement~\cite{Landis}. There are only 6 cases with maximum disagreement (7 experts put a different answer on the first place). We invited an independent physician to analyze these cases for the cause of such inconsistency. The physician concluded that all the answers can be treated as relevant given the provided information. The only difference come from diagnosis ordering: each expert select reasonable diagnosis set, but order them based on their primary specialization, experience and other subjective factors that influence the decision making. We assessed Hit@3 metric for each of the 7 experts by inferring the ground truth diagnoses from the panel of 6 experts' answers and comparing them with the remaining expert answers (thus deploying one-against-all labelling strategy~\cite{Bishop}). It turned out that the Hit@3 metric varied quite a bit across the clinicians. The worst performing expert reached the value of 57.89\% and the best one reached the value of 72.52\%, with the mean Hit@3 value for the experts being 68.16\% and standard deviation of 4.82\%. Our best performing DL-based model (RuPool-BERT 265) against the same 7 panels achieved mean Hit@3 value of 69.15\% with standard deviation of 8.52\%. To test whether the difference between Hit@3 performance metrics of the models and the experts are significant, we applied the Mann–Whitney U test. Our null hypothesis is that the expert and the model metrics are the same, and the alternative is that there is a significant difference in the metrics. The computed empirical value is \(U\textsubscript{emp}=22\) and the critical value for the sample size \(n=7\) at the 0.05 significance level is \(U\textsubscript{crit}=11\). Thus, we fail to reject the null hypothesis and cannot claim that our model is better than humans. Further, based on the model and expert mean metrics, we can conclude that our proposed model shows the performance results comparable to the performance of the medical experts.

\section{Conclusions}
In this paper we described the challenging and important problem of diagnoses prediction from unstructured real-world clinical text data based on a very large Russian EHR dataset containing about 4 million doctor's visits of over 1 million patients. To provide the diagnosis, we proposed a novel BERT-based model for classification of textual clinical notes, called RuPool-BERT, that differs from the others BERT-based approaches by introducing a novel way of the FC-layer composition. Our experiments of applying the developed prediction model to the practical task of classifying 265 diseases showed the advantage of this model compared to the fine-tuned RuBERT base analog and other text representation models. We also showed that using a BERT model with a vocabulary and pretraining dataset tailored to the medical texts representation (RuEHR-BERT) improves performance on the classification task, specially on less frequent diseases. This improvement is achieved at a small fraction of pretraining time compared to the general Russian language model (2 weeks of Tesla K40 for RuEHR-BERT vs 8 weeks of Tesla P100 for RuBERT).

Comparison of our model with a panel of medical experts showed that the results of our model were similar to the results of experts in terms of the Hit@3 performance measure. Furthermore, we showed that the most reliable performance of our system is achieved on those samples having longer textual inputs, i.e. text sequences having at least 20 input tokens. All this allows us to conclude that our model and system has a strong potential to help doctors with disease diagnosis by providing the "second opinions" to them.

Our partners in the medical community identified one issue with the proposed method: they maintain that the proposed approach would benefit greatly from clear explanations of how our method arrived at each particular diagnoses. This is the topic of future research on which we plan to focus in the immediate future.

%
%
%

\begin{thebibliography}{8}
\bibitem{ref1}
Atasoy, H., Greenwood, B. N., McCullough J. S.: The digitization of patient care: A review of the effects of electronic health records on health care quality and utilization, Annual Review of Public Health, No. 40 (5) (2019), pp.~487–500.

\bibitem{ref2}
Fast Healthcare Interoperability Resources (FHIR), \href{https://www.hl7.org/fhir/}{https://www.hl7.org/fhir/}, last accessed 2020/04/20.

\bibitem{ref6}
Liberman, A. L., Newman-Toker, D. E.: Symptom-Disease Pair Analysis of Diagnostic Error (SPADE): a conceptual framework and methodological approach for unearthing misdiagnosis-related harms using big data, BMJ Qual Saf, vol. 27 (2018), No. 7, pp.~557–566.

\bibitem{ref7}
Vardanyan, G.J., Avetisyan, G.A., Janoyan, G.J., Porksheyan, K.A., Khachatryan, A.R., Edilyan, A.R., Minasyan, I.S.: Medical Errors: Modern Condition of the Problem, Medical Science of Armenia, No. 59(4) (2019) pp.~105–120.

\bibitem{icd10}
World Health Organization.: International Statistical Classification of Diseases and Related Health Problems. 10th revision, Fifth edition, 2016.

\bibitem{ref20}
Kuratov, Y., Arkhipov, M.: Adaptation of Deep Bidirectional Multilingual Transformers for Russian Language. arXiv preprint arXiv:1905.07213, 2019.

\bibitem{ref13}
Vasiljeva, I., Arandjelovic, O.: Diagnosis prediction from electronic health records (EHR) using the binary diagnosis history vector representation, Journal of Computational Biology 24(8) (2017), pp.~767–786.

\bibitem{ref14}
Ma, F., Chitta, R., Zhou, J., You, Q., Sun, T., Gao, J.: Dipole: diagnosis prediction in healthcare via attention-based bidirectional recurrent neural networks, Proceedings of the 23rd ACM SIGKDD International Conference on Knowledge Discovery and Data Mining (2017), pp.~1903–11, \doi{10.1145/3097983.3098088}.

\bibitem{ref15}
Shi, J., Fan, X., Wu, J., et al.: DeepDiagnosis: DNN-Based Diagnosis Prediction from Pediatric Big Healthcare Data, Sixth International Conference on Advanced Cloud and Big Data (CBD) IEEE (2018), pp.~287–292.

\bibitem{ref8}
Qiao, Z., Wu, X., Ge, S., Fan, W.: Mnn: multimodal attentional neural networks for diagnosis prediction, Proceedings of the 28th International Joint Conference on Artifcial Intelligence, AAAI Press, (2019), pp.~5937–5943.

\bibitem{ref16}
Devlin, J., Chang, M.-W., Lee, K., et al.: Bert: Pretraining of deep bidirectional transformers for language understanding. arXiv preprint arXiv:1810.04805, 2018.

\bibitem{ref17}
Amin, S., Neumann, G., Dunfield, K., Vechkaeva, A., Chapman, K., Wixted, M.: MLT-DFKI at CLEF eHealth 2019: Multi-label Classification of ICD-10 Codes with BERT, Conference and Labs of the Evaluation Forum (Working Notes) (2019).

\bibitem{ref18}
Li, F., Jin, Y., Liu, W., et al.: Fine-Tuning Bidirectional Encoder Representations From Transformers (BERT)–Based Models on Large-Scale Electronic Health Record Notes: An Empirical Study, JMIR medical informatics, 7(3), (2019), \doi{10.2196/14830}.

\bibitem{ref12}
Sakhibgareeva, M.V., Zaozersky, A.Y.: Developing an artificial intelligence-based system for medical prediction, Bulletin of Russian State Medical University, No 6 (2017), pp. 42-46, \doi{10.24075/brsmu.2017-06-07}.

\bibitem{ref11}
Johnson, A.E.W., et al.: MIMIC-III, a freely accessible critical care database. Scientific Data 3: 160035 (2016), \doi{10.1038/sdata.2016.35}.

\bibitem{ref9}
Ma, F., Chitta, R., Zhou, J., You, Q., Sun, T., Gao, J.: Dipole: diagnosis prediction in healthcare via attention-based bidirectional recurrent neural networks, Proceedings of the 23rd ACM SIGKDD International Conference on Knowledge Discovery and Data Mining (2017), pp.~1903–11, \doi{10.1145/3097983.3098088}.

\bibitem{ref10}
Malakouti, S., Hauskrecht, M.: Predicting patient's diagnoses and diagnostic categories from clinical-events in EHR data, Conference on Artificial Intelligence in Medicine, (2019), pp.~125–130, \doi{10.1007/978-3-030-21642-9\_17}.

\bibitem{ref19}
Wu, Y., Schuster, M., Chen, Z., et al.: Google’s neural machine translation system: Bridging the gap between human and machine translation. arXiv preprint arXiv:1609.08144, 2016.

\bibitem{ref21}
Bojanowski, P., Grave, E., Joulin, A., Mikolov, T.: Enriching word vectors with subword information, Transactions of the Association for Computational Linguistics, No. 5 (2017), pp.~135–146.

\bibitem{ref22}
Cer, D., Yang, Y., et al.: Universal sentence encoder, CoRR. arXiv preprint arXiv:1803.11175, 2018.

\bibitem{Manning}
Manning, C., Raghavan, P., and Schutze H.: Introduction to information retrieval. Cambridge University Press, (2008).

\bibitem{Landis}
Landis, J. R. and Koch, G. G.: The measurement of observer agreement for categorical data, Biometrics, vol. 33 (1977), No. 1, pp.~159–174.

\bibitem{Bishop}
Bishop, C. M.: Pattern Recognition and Machine Learning. Springer, (2006).

\end{thebibliography}
%

\end{document}